\def\year{2022}\relax
\documentclass[letterpaper]{article} % DO NOT CHANGE THIS
\usepackage{aaai22}  % DO NOT CHANGE THIS
\usepackage{times}  % DO NOT CHANGE THIS
\usepackage{helvet}  % DO NOT CHANGE THIS
\usepackage{courier}  % DO NOT CHANGE THIS
\usepackage[hyphens]{url}  % DO NOT CHANGE THIS
\usepackage{graphicx} % DO NOT CHANGE THIS
\usepackage{natbib}  % DO NOT CHANGE THIS AND DO NOT ADD ANY OPTIONS TO IT
\usepackage{caption} % DO NOT CHANGE THIS AND DO NOT ADD ANY OPTIONS TO IT
\DeclareCaptionStyle{ruled}{labelfont=normalfont,labelsep=colon,strut=off} % DO NOT CHANGE THIS
\usepackage[dvipsnames]{xcolor}         % colors
\usepackage{amsmath}
\usepackage{algorithm}
\usepackage{algorithmic}
\usepackage{newfloat}
\usepackage{multirow}
\usepackage{amsfonts}
\usepackage{subcaption}
\usepackage{booktabs}
\usepackage{microtype}
\usepackage{nicefrac}
\usepackage[misc]{ifsym}
\usepackage{listings}
\usepackage{tikz}
\newcommand*\circled[1]{\tikz[baseline=(char.base)]{
            \node[shape=circle,draw,inner sep=0.2pt] (char) {#1};}}
\floatstyle{ruled}
\newfloat{listing}{tb}{lst}{}
\floatname{listing}{Listing}
\usepackage{multirow}
\usepackage{xspace}
\usepackage[dvipsnames]{xcolor}
\newcommand{\ourapproach}{\textsc{DANet}\xspace}
\newcommand{\ourcomponent}{\textsc{AbstLay}\xspace}
\usepackage{comment}
\title{{\ourapproach}s: Deep Abstract Networks for Tabular Data Classification and Regression}
\author{
    Jintai Chen\textsuperscript{\rm 1},
    Kuanlun Liao\textsuperscript{\rm 1},
    Yao Wan\textsuperscript{\rm 2},
    Danny Z. Chen\textsuperscript{\rm 3},
    Jian Wu\textsuperscript{\rm 4,}\footnote{The corresponding author.}
}
\affiliations{
  \textsuperscript{\rm 1} College of Computer Science and Technology, Zhejiang University, Hangzhou, China\\
  \textsuperscript{\rm 2} School of Computer Science and Technology, Huazhong University of Science and Technology, Wuhan, China\\
  \textsuperscript{\rm 3} Department of Computer Science and Engineering, University of Notre Dame, Notre Dame, IN 46556, USA\\
  \textsuperscript{\rm 4} The First Affiliated Hospital, and Department of Public Health, Zhejiang University School of Medicine, Hangzhou, China\\
  jtchen721@gmail.com, stevekll@zju.edu.cn, wanyao@hust.edu.cn, dchen@nd.edu, wujian2000@zju.edu.cn
}

\usepackage{bibentry}

\begin{document}

\maketitle

\begin{abstract}
Tabular data are ubiquitous in real world applications. Although many commonly-used neural components (e.g., convolution) and extensible neural networks (e.g., ResNet) have been developed by the machine learning community, few of them were effective for tabular data and few designs were adequately tailored for tabular data structures. In this paper, we propose a novel and flexible neural component for tabular data, called \textit{Abstract Layer} (\ourcomponent), which learns to explicitly group correlative input features and generate higher-level features for semantics abstraction. Also, we design a structure re-parameterization method to compress the trained \ourcomponent, thus reducing the computational complexity by a clear margin in the reference phase.
A special basic block is built using {\ourcomponent}s, and we construct a family of \textit{\textbf{D}eep \textbf{A}bstract \textbf{Net}works} ({\ourapproach}s) for tabular data classification and regression by stacking such blocks. In {\ourapproach}s, a special shortcut path is introduced to fetch information from raw tabular features, assisting feature interactions across different levels. Comprehensive experiments on seven real-world tabular datasets show that our \ourcomponent and {\ourapproach}s are effective for tabular data classification and regression, and the computational complexity is superior to competitive methods. Besides, we evaluate the performance gains of {\ourapproach} as it goes deep, verifying the extendibility of our method. Our code is available at \url{https://github.com/WhatAShot/DANet}.
\end{abstract}

\section{Introduction}\label{sec:intro}
Data organized in tabular structures, e.g., medical indicators~\cite{hassan2020machine,mirroshandel2016applying} and banking records~\cite{roy2018deep,babaev2019rnn,addo2018credit}, are ubiquitous in daily life. However, unlike the boom of deep learning in the computer vision and natural language processing fields, very few neural networks were adequately designed for tabular data~\cite{tabnet,yang2018deep,tabnn,roy2018deep,babaev2019rnn,FCNN,guo2017deepfm}, and hence the performances (e.g., in classification and regression tasks) of such neural networks were still somewhat inferior~\cite{dnfnet}.
Inspired by the success of ensemble learning (e.g., XGBoost)~\cite{gbdt,XGBoost,lightbgm,CatBoost,randomforest} on tabular data, some recent work resorted to combining multiple neural networks within the framework of ensemble learning~\cite{NODE,dnfnet,deepgbm}.
Although ensemble learning can boost the performances of neural networks on tabular data (in the cost of increased computational resources), with such methods, the power of neural networks in tabular feature processing is not yet fully exploited. Besides, there are not many efficient neural components specifically designed for tabular data (analogous to convolution for computer vision). Consequently, known neural networks were mainly based on sundry components and thus were not very extensible.
\begin{figure*}[t]
    \centering
    \includegraphics[width=0.95\textwidth]{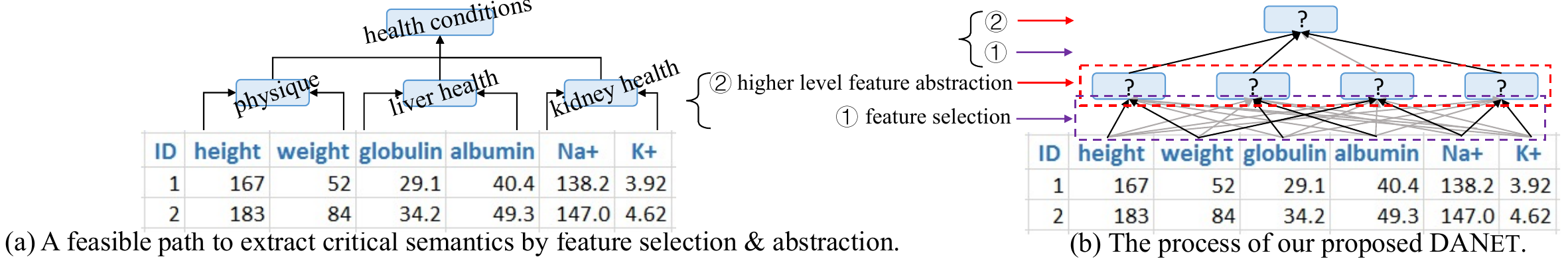}
    \caption{\textbf{A running example of health assessment for illustrating our insights.} 
    (a) A feasible semantics-oriented feature abstraction process. There are three underlying feature groups that can be found to compute high-level features measuring \textit{physique}, \textit{liver health}, and \textit{kidney health}; then these three features are further grouped to estimate the health conditions.
    (b) An \ourcomponent \circled{1}~learns a proper feature selection bias to group correlative features and then \circled{2}~abstracts meaningful higher-level features, and {\ourapproach}s organize {\ourcomponent}s to repeat this process until finally obtaining global semantics for health assessment. The blue rectangles denote the computed high-level features, the grey lines indicate the candidates for feature selection, and the black arrows mark the features eventually selected.}
    \label{fig:motivation}
\end{figure*}

In this paper, we present a flexible neural component called \textit{Abstract Layer} (\ourcomponent) for tabular feature abstraction, and build \textit{Deep Abstract Networks} ({\ourapproach}s) based on {\ourcomponent}s for tabular data classification and regression. Since tabular features are generally irregular, it is hard to introduce fixed inductive biases (such as dependency among spatially neighboring features in images) in designing neural networks for tabular data processing (e.g., classification and regression). To this end, we assume that there are some underlying feature groups in a tabular data structure, and the features in the groups are correlative and can be exploited to attain higher-level features relevant to the prediction targets. We propose to decouple the process of higher-level tabular feature abstraction into two steps: (i) correlative feature grouping, and (ii) higher-level feature abstraction from grouped features. We employ an {\ourcomponent} to perform these two steps, and {\ourapproach}s repeat these two steps by stacking {\ourcomponent}s to represent critical semantics of tabular data.

Fig.~\ref{fig:motivation} gives a running example to illustrate our insights. As shown in Fig.~\ref{fig:motivation}(a), feasible underlying feature groups and the potential feature abstraction paths are organized as follows. The \textit{height} and \textit{weight} can be grouped together to compute more comprehensive features that represent the \textit{physique}.
Similarly, features representing \textit{liver health} and \textit{kidney health} can be abstracted from the raw features, and the features representing \textit{health conditions} can further be abstracted from the three high-level features. The semantics are hierarchically aggregated, and the whole process is presented as a parse tree. In contrast, in Fig.~\ref{fig:motivation}(b), our method learns to find and group correlative features and then abstract them into higher-level features. This process repeats until global semantics are obtained. The higher-level tabular features are abstracted by one neural layer (\ourcomponent), and the hierarchical abstraction process is realized with deep learning networks. That is why we call them \textit{Abstract Layer} and \textit{Deep Abstract Networks}, respectively.

In designing \ourcomponent, we contemplate how to group features and abstract them into higher-level features. Since it is hard to find a metric space to measure the feature diversities for feature grouping due to the heterogeneity of tabular data, our \ourcomponent learns to group the features through employing learnable sparse weight masks, without introducing any explicit distance measurement.
Then, subsequent feature learners (in the \ourcomponent) are utilized to abstract higher-level features from the respective feature groups. Further, motivated by the structure re-parameterization~\cite{RepVGG}, we develop a specific re-parameterization method to merge the two step operations of \ourcomponent into one step in the inference phase, reducing the computational complexity.

\begin{figure*}
    \centering
    \includegraphics[width=0.78\textwidth]{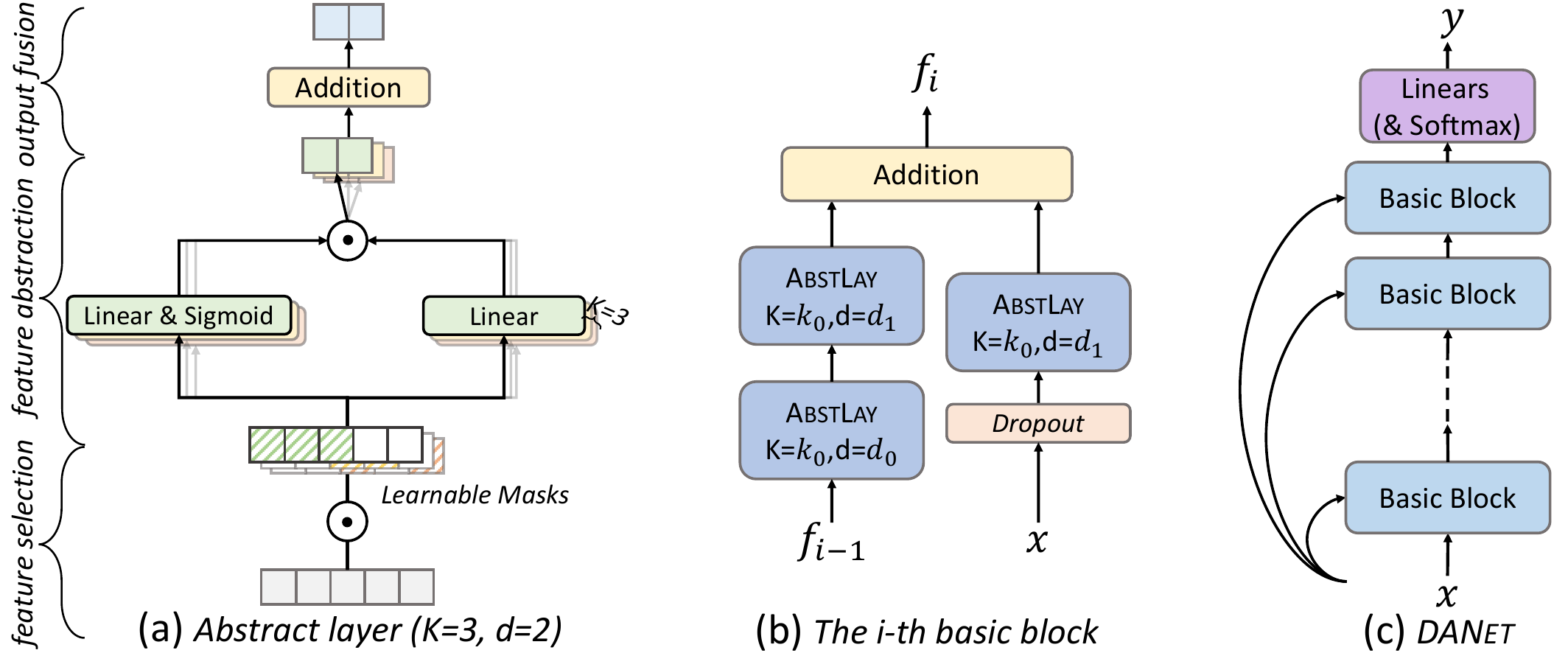}
    % \vskip -0.8em
    \caption{\textbf{Our proposed architecture for tabular data processing.} (a) Illustrating an \ourcomponent, which performs three steps: \textit{feature selection}, \textit{feature abstracting}, and \textit{output fusion}. In the example of (a), the number of masks, $K$, is set to 3 (see Eq.~(\ref{eq:abs})), the output feature dimension, $d$, is set to 2, and $\odot$ indicates the element-wise multiplication. (b) Illustrating a basic block specification. (c) The architecture of {\ourapproach} is built mainly by stacking several basic blocks.}\label{fig:framework}
    % \vskip -0.15 in
\end{figure*}

Our {\ourapproach}s are built mainly by stacking {\ourcomponent}s sequentially, and thus tabular features are recursively abstracted layer by layer to obtain the global semantics. To replenish useful features and increase the feature diversity, we also introduce a shortcut path (similar to the residual shortcut~\cite{resnet}), which directly injects the information of raw tabular features into the higher-level features. Specifically, we package the higher-level feature abstraction operation of \ourcomponent and the feature fetching operation of the shortcut path into a basic block (as specified in Fig.~\ref{fig:framework}(b)), and our {\ourapproach}s are built by stacking such blocks (see Fig.~\ref{fig:framework}(c)).
Note that various empirical evidences~\cite{resnet,Pointnet++} have suggested that the successes of deep neural networks (DNNs) are partially benefited from the model depth. Thus, we design {\ourapproach}s with deep architectures, and further discuss the benefits and choices of the model depths by extensive experiments.

In summary, the contributions of this paper are as follows.
\begin{itemize}
    \item The proposed \ourcomponent automatically extracts higher-level tabular feature from lower-level ones. \ourcomponent is simple, and its computational complexity can be reduced in inference by our structure re-parameterization method.
    \item We introduce a special shortcut path, which fetches raw features for higher levels, promoting the feature diversity for finding meaningful feature groups.
    \item Based on {\ourcomponent}s, we build {\ourapproach}s to cope with tabular data classification and regression tasks by recursively abstracting features in order to obtain critical semantics of tabular features. 
    % {\ourapproach}s outperform previous methods on multiple public datasets.
\end{itemize}
\section{Related Work}
\paragraph{Tabular data processing.}
Various conventional machine learning methods~\cite{he2014practical,cart,XGBoost,zhang2006learning,zhang2003learning} were proposed for tabular data classification and learn-to-rank (regression). Decision tree models~\cite{ID3,cart} can present clear decision paths and are robust on simple tabular datasets. Ensemble models based on decision trees, such as GBDT~\cite{gbdt}, LightGBM~\cite{lightbgm}, XGBoost~\cite{XGBoost}, and CatBoost~\cite{CatBoost}, are currently top choices for tabular data processing, and their performances were comparable~\cite{anghel2018benchmarking}. 

Currently, a research trend aimed to apply DNNs~\cite{guo2017deepfm,yang2018deep} onto tabular datasets. Some neural networks under the ensemble learning frameworks were presented in~\cite{lay2018random,feng2018multi}. Recently, NODE~\cite{NODE} combined neural oblivious decision trees with dense connections and obtained comparable performances as GBDTs~\cite{gbdt}. Net-DNF~\cite{dnfnet} implemented soft versions of logical boolean formulas to aggregate the results of a large number of shallow fully-connected models. Both NODE and Net-DNF essentially followed ensemble learning, employing many (e.g., 2048) shallow neural networks, and thus were computing-complex. Such strategies did not explore the potential of deep models, and their performances should be attributed largely to the number of sub-networks. TabNet~\cite{tabnet} computed sparse attentions sequentially to imitate the sequential feature splitting procedure of tree models. However, TabNet was verified to attain slightly inferior performances, as noted in~\cite{dnfnet}.
\paragraph{Feature selection.}
Since  tabular features are heterogeneous and irregular, various feature selection methods were applied previously. Classical tree models often used information metrics to guide feature selection, such as information gain~\cite{ID3}, information gain ratio~\cite{c45}, and Gini index~\cite{cart}, which are essentially greedy algorithms and may require branch pruning~\cite{c45} or early stopping strategy. Decision tree ensemble methods often applied random feature sampling to promote diversity. To further assist feature selection, some bagging methods utilized the out-of-bag estimate~\cite{james2013introduction}, and gcForest~\cite{deepforest} used sliding windows to scan and group raw features for different forests. A fully-connected neural network~\cite{FCNN} blindly took in all the features, and TabNN~\cite{tabnn} selected features based on ``data structural knowledge'' learned by GBDTs. Most of tree models selected one single feature in a step, ignoring the underlying feature correlations.

At present, some neural networks introduced neural operations to select features. NODE~\cite{NODE} employed learnable feature selection matrices with Heaviside functions for hard feature selection, imitating the processing of oblivious decision trees. A key to NODE is that the back-propagation optimization is used to replace the information metrics in training the ``tree'' models. However, the parameters specified by Heaviside functions are hard to update via back-propagation, and thus NODE may take many iterations before convergence. Net-DNF used a straight-through trick~\cite{bengio2013estimating} to optimize this issue, but it required extra loss functions in training feature selection masks and was inconvenient for users. TabNet~\cite{tabnet} employed an attention mechanism for feature selection, but selected different features for different instances; hence, it is difficult to capture stable feature correlations. In contrast, this paper seeks to find the underlying feature groups representing target-relevant semantics and develop the corresponding operations that are simple and user-friendly.

\section{Problem Statement}
Suppose $\mathcal{X}=(\mathcal{F}, X, y)$ is one type of specific tabular data structure, where $\mathcal{F}$ specifies the raw feature type space, $X$ is the feasible instance space, and $y$ is the target space. In a tabular dataset of the $\mathcal{X}$ type, an instance $x \in \mathbb{R}^n$ in $X$ is defined as an $n$-element vector representing $n$ scalar raw features in $\mathcal{F}$ ($n=\lvert\mathcal{F}\rvert$). Notably, tabular data features are irregular, and the feature permutation in $x$ is predefined. In this paper, we assume that there are some underlying feature groups in a tabular data structure, and the features in a group are correlative and target-relevant. Note that some features may be in no group and some in multiple groups. We are interested in learning mapping functions that take $x \in X$ as input, dig out and address the underlying feature groups for target semantic interests (determined by the classification/regression tasks).

\section{Abstract Layer}\label{sec:abslay}
\subsection{Key Functions and Operation}
We propose an \textit{Abstract layer} (\ourcomponent), which learns to find some underlying feature groups and abstract higher-level features by processing the grouped features. The \ourcomponent is also desired to be flexible and simple as a basic layer.
% Thus the methods requiring additional supervisions such as TabNet and Net-DNF are inadequate.
In our design, the \ourcomponent comprises \textit{feature selection} functions to find feature groups, subsequent \textit{feature abstracting} functions to abstract higher-level features from groups, and an \textit{output fusion} operation to fuse features abstracted from various groups, as shown in Fig.~\ref{fig:framework}(a).
\paragraph{Feature selection function.} Given an input vector $f \in \mathbb{R}^m$ containing $m$ scalar features, a learnable sparse mask $M\in \mathbb{R}^m$ selects a subset of scalar features from $f$ for one group. Specifically, this learnable mask is defined as a learnable parameter vector $W_{\text{mask}}$ followed by the Entmax sparsity mapping~\cite{Entmax}, and the features are selected by element-wise multiplying with the sparse mask $M$. The Entmax is a variational form~\cite{wainwright2008graphical} of the Softmax, which introduces sparsity to the output probability. Formally, the \textit{feature selection} is defined by
\begin{equation}\label{eq:mask}
    M = \text{entmax}_\alpha(W_{\text{mask}})\ , \quad f^\prime = M \odot f \ ,
\end{equation}
where the parameter vector $W_{\text{mask}} \in \mathbb{R}^m$, $\odot$ denotes element-wise multiplication, and the selected features are presented in $f^\prime \in \mathbb{R}^m$. In the Entmax sparsity mapping, we use the default setting with $\alpha=1.5$. With the multiplication, there are some zero values in $f^\prime$, and a zero value for the $i$-th scalar feature of the vector $f^\prime$ means that the $i$-th scalar feature in $f$ is not selected. This feature selection is simple and can select identical features for different instances.

\paragraph{Feature abstracting function.} Given the selected features in $f^\prime \in \mathbb{R}^{m}$ (as defined above), we define the \textit{feature abstracting} function using a fully connected layer with a simple attention mechanism~\cite{GLU}. Formally, the output $f^*$ of a \textit{feature abstracting} function is computed by
\begin{equation}\label{eq:processing}
    q = \text{sigmoid}(\text{BN}(W_1 f^\prime))\ , \quad f^* = \text{ReLU}(q \odot \text{BN}(W_2 f^\prime))\ ,
\end{equation}
where the two learnable parameters $W_c \in \mathbb{R}^{d \times m}$ ($c=1, 2$) are equal-sized, and $q$ denotes the computed attention vector. $W_c f^\prime$ were implemented by 1D convolutions, and the parametric biases were ignored in Eq.(\ref{eq:processing}).
Since tabular data are often trained with a large batch size, we use the ghost batch normalization~\cite{GBN} to operate ``BN''. In this way, the selected features in the vector $f^\prime \in \mathbb{R}^m$ are projected to $f^* \in \mathbb{R}^d$, and we treat the $d$ values in the feature vector $f^*$ as independent scalar features representing various semantics. Note that all the $d$ features of $f^*$ are abstracted from the same group (determined by the same $M$ in Eq.~(\ref{eq:mask})).

\paragraph{Parallel processing and output fusion.} The effect of the \ourcomponent is realized primarily by the \textit{feature selection} function and \textit{feature abstracting} function. These two functions work in sequence to abstract higher-level features from the lower-level feature groups. Yet, we consider that more than one group can be found in a given feature vector $f$. Also, it is common that informative output features are typically obtained by applying some unit operations in parallel (e.g., a convolution layer often contains many kernels). Motivated by these, our \ourcomponent is designed to find and process multiple low-level feature groups in parallel. Formally, we specify its computation by
\begin{equation}\label{eq:abs}
    f_{o} = \sum^K_{k=1} p_k \circ s_k(f)\ ,
\end{equation}
where $p \ \circ \ s$ denotes the composite function of a \textit{feature selection} function $s$ and a \textit{feature abstracting} function $p$, and $K$ is the number of feature groups that \ourcomponent manages to get and is a hyper-parameter. We set the output feature sizes of all $p_k \circ s_k$ identical. The output features of all the composite functions $p_k \circ s_k$ are element-wise added to form the output features $f_o$ of \ourcomponent (see Fig.~\ref{fig:framework}(a)). 

Similar to the convolution layers in a model, several {\ourcomponent}s can be stacked together and operate as a whole. Thus, the output scalar features of one \ourcomponent may be further grouped by its subsequent \ourcomponent for further information abstraction, and the useless output features from the preceding \ourcomponent can be abandoned. Different from the complicated ``feature transformation'' function in TabNet~\cite{tabnet}, the ability of the {\ourcomponent}s is largely due to their co-operation (e.g., layer-by-layer processing).

\subsection{\ourcomponent Complexity Reduction}\label{sec:ess}
To reduce the computational complexity of our proposed \ourcomponent, we develop a re-parameterization method following~\cite{RepVGG} to reformulate the {\ourcomponent}s.
Note that $W_1 \in \mathbb{R}^{d \times m}$ and $W_2 \in \mathbb{R}^{d \times m}$ are weights of \textit{feature abstracting function}, and $M \in \mathbb{R}^m$ is also a weight vector. Substituting Eq.~(\ref{eq:mask}) into Eq.~(\ref{eq:processing}), we have
\begin{equation}\label{eq:substitute}
\begin{aligned}
    & q = \text{sigmoid}(\text{BN}(W_1 (M\odot f)))\ , \\
    & f^* = \text{ReLU}(q \odot \text{BN}(W_2 (M\odot f)))\ .
\end{aligned}
\end{equation}
Thus, we can use $W_c^\prime \in \mathbb{R}^{d\times m}$ to replace the multiplication term $W_c M$ ($c=1, 2$) in Eq.~(\ref{eq:substitute}), by
\begin{equation*}
    W_c^\prime[:,j] = W_c[:,j] * M[j]\ ,
\end{equation*}
where $j=1, 2, \ldots, m$, and $m$ is the input feature dimension. Besides, we can further merge the batch normalization operation into the weights $W^\prime_c$, by
% and obtain $W^*_c$, by
\begin{equation}\label{eq:bn_in}
\begin{aligned}
    & W^*_c[i,:] = \frac{\gamma[i]}{\sigma[i]} W_c^\prime[i,:]\ ,
    & b^*_c[i] = (\beta[i] - \frac{\mu[i] \gamma[i]}{\sigma[i]})\ ,
    \end{aligned}
\end{equation}
where $i=1, 2, \ldots, d$ and $d$ is the output feature dimension, $\gamma\in \mathbb{R}^d$ and $\beta\in \mathbb{R}^d$ are the learnable parameters of the batch normalization followed $W_c$ (the formula is $z^\prime = \frac{\gamma}{\sigma} z + (\beta - \frac{\mu \gamma}{\sigma})$ for a feature vector $z$), and $\mu \in \mathbb{R}^d$ and $\sigma \in \mathbb{R}^d$ are the computed mean and standard deviation. Then, the operation in an \ourcomponent (see Eq.~(\ref{eq:abs})) can be simplified as
\begin{equation}\label{simplify}
    f_o = \sum^K_{k=1} \text{ReLU}(\text{sigmoid}(W^*_{k,1} f + b^*_{k,1}) \odot (W^*_{k,2} f + b^*_{k,2}))\ ,
\end{equation}
where $W^*_{k,c}$ ($c=1, 2$) is the weights $W^*_{c}$ re-parameterized by Eq.~(\ref{eq:bn_in}) for the $k$-th \textit{feature abstracting} function in an \ourcomponent (see Eq.~(\ref{eq:abs}), an \ourcomponent has $K$ functions), and $b^*_{k,c}$ is the $b^*_c$ in Eq.(\ref{eq:bn_in}) for the $k$-th \textit{feature abstracting} function. In this way, a lighter model can be used in inference by re-parameterization.

\section{Deep Abstract Networks}
Based on the proposed \ourcomponent, we introduce \textit{\textbf{D}eep \textbf{A}bstract \textbf{Net}works} ({\ourapproach}s) for tabular data processing.
% As mentioned in Sec.~\ref{sec:intro}, artificial feature engineering can recursively group features and abstract higher-level features. Motivated by this, 
{\ourapproach}s stack {\ourcomponent}s to repeatedly find and process some meaningful feature groups for higher-level feature abstraction. Besides, we allow features in different levels to be grouped together, thus increasing the model capability. Hence, we design a new shortcut path that allows a high-level layer to fetch raw features. Specifically, we propose a basic block based on {\ourcomponent}s containing the shortcut path, and our {\ourapproach}s are built by sequentially stacking such blocks. 
% as shown in Fig.~\ref{fig:framework}(c).

\subsection{A Basic Block}\label{sec:block}
Our basic block is mainly built using {\ourcomponent}s, and a new shortcut can add features abstracted from the groups of raw features to the main model path. Fig.~\ref{fig:framework}(b) illustrates the specification of the basic block in {\ourapproach}s. Formally, we define the $i$-th basic block $f_{i}$ by
\begin{equation}\label{eq:shortcut}
    f_{i} = \mathcal{G}_i(f_{i-1}) + g_i(x)\ ,
\end{equation}
where $g_i$ is the shortcut consists of an \ourcomponent and a Dropout layer~\cite{dropout} and takes raw features $x$ as input. The term $\mathcal{G}_i$ is on the main path containing multiple {\ourcomponent}s and its input is the features produced by the previous basic block (see Fig.~\ref{fig:framework}(c)). For the first basic block $f_1$, we let $f_{0}=x$. Unlike the residual block in ResNet~\cite{resnet} whose shortcut path brings the features of the preceding layers, our shortcut fetches the raw features.

In a {\ourapproach} with many basic blocks (see Fig.~\ref{fig:framework}(c)), the combination of $\mathcal{G}_i$'s of the basic blocks acts as the main path of the model, which extracts and forwards target-relevant information. The target-relevant information is replenished continuously via the shortcut terms $g_i$ of the basic blocks. From Fig.~\ref{fig:framework}(c), it is obvious that a raw feature can be used by a high-level basic block via a shortcut directly, while the information of some raw features may be taken by a higher-level layer through the main path after the layer-by-layer processing. Thus, the feature diversity in a layer increases compared to a layer in a model without such shortcuts. Notably, we include a Dropout operation in the shortcut path, which encourages the subsequent \ourcomponent to focus on the core information that the basic block requires.
\begin{table*}[th]
\centering
\caption{\textbf{A summary of the seven public datasets.} The datasets marked with ``\dag'' are randomly split into training and test sets by a ratio of 8:2. (``Forest'': ``Forest Cover Type''; ``Cardio.'': ``Cardiovascular Disease''; ``L2R'': ``Learn to Rank''; ``Clas.'': ``Classification''.)}\label{tab:data}
% \small
\begin{tabular}{l|ccccccc}
\toprule
\multicolumn{1}{c|}{Datasets} & YearPrediction & Microsoft & Yahoo & Epsilon & Click & Cardio.\dag & Forest\dag \\ \midrule
\# Features  & 90   & 136  & 699  & 2K  & 11  & 11 & 54  \\ \midrule
Size of train data  & 463K  & 723K  & 544K & 400K & 900K & 56K & 400K \\ \midrule
Size of test data  & 51.6K & 241K  & 165K &  100K & 100K & 14K & 100K \\ \midrule
Task types & L2R & L2R & L2R & Clas. & Clas. & Clas. & Clas. \\ \midrule
Metric  & MSE  & MSE  & MSE  & Acc.  & Acc. & Acc.   & Acc.  \\ \bottomrule
\end{tabular}
\end{table*}
\begin{table*}[t]
\caption{\textbf{Performance comparison on the seven tabular datasets.} The best performances are marked in \textcolor{Peach}{orange}, and the second and third best ones are marked in \textcolor{RoyalBlue}{blue} and \textcolor{ForestGreen}{green}, respectively. Note that for classification tasks, a better method gets a higher accuracy, and for learn-to-rank tasks, a better method gets a lower MSE.}
% Some of performances of the compared methods are from their original papers or~\cite{NODE}.}
\label{tab:performances}
% \vskip -0.15 in
\setlength{\tabcolsep}{2pt}
\centering
\small
\begin{tabular}{l|c|cccc|ccc}
\toprule
% \multicolumn{7}{c}{Datasets}
\multicolumn{1}{c|}{\multirow{2}{*}{Methods}}&  \multirow{2}{*}{Rank} & \multicolumn{4}{c|}{Classification} & \multicolumn{3}{c}{Learn-to-rank} \\ \cline{3-9}
 & & Forest  & Cardio. & Epsilon & Click & Microsoft  & YearP. & Yahoo  \\ \midrule
XGBoost~\cite{XGBoost} & 4 & {97.13\%}\tiny{$\pm$2e-4}  &  73.97\%\tiny{$\pm$2e-4}  &  88.89\%\tiny{$\pm$6e-4}  &  66.66\%\tiny{$\pm$2e-3} &  \textcolor{Peach}{0.5544}\tiny{$\pm$1e-4} &  78.53\tiny{$\pm$0.09}  &  \textcolor{Peach}{0.5420}\tiny{$\pm$4e-4}\\
CatBoost~\cite{CatBoost} & 5 &  95.67\%\tiny{$\pm$4e-4} &  \textcolor{RoyalBlue}{74.02\%}\tiny{$\pm$1e-4} & 88.87\%\tiny{$\pm$4e-4} &  65.99\%\tiny{$\pm$2e-3} &  0.5565\tiny{$\pm$2e-4} & 79.67\tiny{$\pm$0.12}   &  \textcolor{RoyalBlue}{0.5632}\tiny{$\pm$3e-4} \\ 
gcForest~\cite{deepforest} & -- &  96.29\%  & 73.27\%  &  88.21\%   &  66.67\% &   --  &  --  &  --  \\  \midrule
Net-DNF~\cite{dnfnet} & -- & \textcolor{ForestGreen}{97.21\%}\tiny{$\pm$2e-4} &  73.75\%\tiny{$\pm$2e-4} & 88.23\%\tiny{$\pm$3e-4} & 66.94\%\tiny{$\pm$4e-4} & -- & -- & -- \\
TabNet~\cite{tabnet} & 7 & 96.99\%\tiny{$\pm$8e-4} &  73.70\%\tiny{$\pm$6e-4} &  \textcolor{ForestGreen}{89.65\%}\tiny{$\pm$8e-5} &  66.84\%\tiny{$\pm$2e-4} & 0.5707\tiny{$\pm$3e-4} &  77.36\tiny{$\pm$0.37} & 0.5925\tiny{$\pm$1e-3}\\
NODE~\cite{NODE} & 3 & 96.95\%\tiny{$\pm$3e-4} &  73.93\%\tiny{$\pm$7e-4} &  \textcolor{RoyalBlue}{89.66\%}\tiny{$\pm$3e-4}  &  66.88\%\tiny{$\pm$2e-3} & 0.5570\tiny{$\pm$2e-4}  &\textcolor{RoyalBlue}{76.21}\tiny{$\pm$0.12} &  0.5692\tiny{$\pm$2e-4} \\
FCNN~\cite{FCNN} & 8 & 96.83\%\tiny{$\pm$1e-4}  &  73.86\%\tiny{$\pm$4e-4} &  89.59\%\tiny{$\pm$2e-4} & 66.75\%\tiny{$\pm$2e-3}  &  0.5608\tiny{$\pm$4e-4} &  79.99\tiny{$\pm$0.47} &  0.5773\tiny{$\pm$1e-3}\\
FCNN + $l^1$-norm & 5 & 96.85\%\tiny{$\pm$1e-3} & 73.90\%\tiny{$\pm$5e-4} &  89.49\%\tiny{$\pm$2e-3} &  \textcolor{ForestGreen}{67.01\%}\tiny{$\pm$2e-4}  &  0.5694\tiny{$\pm$1e-3} & \textcolor{ForestGreen}{76.52}\tiny{$\pm$0.02}  & 0.6016\tiny{$\pm$1e-3}\\ \midrule
{\ourapproach}-20 (ours)&  2  & \textcolor{RoyalBlue}{97.23\%}\tiny{$\pm$2e-4} & \textcolor{Peach}{74.04\%}\tiny{$\pm$5e-4}  & 89.58\%\tiny{$\pm$4e-4} & \textcolor{RoyalBlue}{67.11\%}\tiny{$\pm$2e-4}&   \textcolor{RoyalBlue}{0.5550}\tiny{$\pm$7e-4}  & 76.76\tiny{$\pm$0.15} & \textcolor{ForestGreen}{0.5678}\tiny{$\pm$4e-4} \\
{\ourapproach}-32 (ours) &  1 & \textcolor{Peach}{97.27\%}\tiny{$\pm$5e-4} & \textcolor{ForestGreen}{73.98\%}\tiny{$\pm$2e-4} & \textcolor{Peach}{89.67\%}\tiny{$\pm$2e-4} &  \textcolor{Peach}{67.19\%}\tiny{$\pm$5e-4}  &  \textcolor{ForestGreen}{0.5557}\tiny{$\pm$3e-4} &  \textcolor{Peach}{75.93}\tiny{$\pm$0.17}  & 0.5703\tiny{$\pm$6e-5}  \\
\bottomrule
\end{tabular}
\vskip -0.1 in
\end{table*}
\subsection{Network Architectures and Training}\label{sec:dan}
We stack the basic blocks in sequence to build a {\ourapproach} architecture, as shown in Fig.~\ref{fig:framework}(c). In our setting, we fix the basic block specification that contains three {\ourcomponent}s, as shown in Fig.~\ref{fig:framework}(b). That is, in Eq.~(\ref{eq:shortcut}), $\mathcal{G}_i$ is composed of two {\ourcomponent}s, and $g_i$ contains one. Then, a three-layer MLP (a multi-layer perceptron network) with ReLU activation is used at the end of a {\ourapproach} for classification (with Softmax) or regression. We have tested various network architecture specifications, and observed consistent patterns. Here, we present some concrete architectures\footnote{The postfix numbers indicate the numbers of {\ourcomponent}s stacked in the main path.}, such as {\ourapproach}-20 and {\ourapproach}-32, to analyze the effects of {\ourapproach}s.

Similar to the previous DNNs for tabular data~\cite{tabnet,NODE}, our \textit{\ourapproach}s can deal with classification and learn-to-rank (regression) tasks on tabular data. \textit{\ourapproach}s are trained with the specification of the Cross-Entropy loss function for classification, and are trained with the mean squared error (MSE) for regression. Note that the feature names are not used in this paper.

\section{Experiments}
In this section, we present extensive experiments to compare the effects of our {\ourapproach}s and the known state-of-the-art models. Also, we present several empirical studies to analyze the effects of some critical {\ourapproach} components, including the learnable sparse masks, shortcut paths, model depth, and model width (the $K$ value in Eq.~(\ref{eq:abs})). Besides, we evaluate the effects of our proposed sparse masks on correlative feature grouping using three synthesized datasets.
\subsection{Experimental Setup}
\paragraph{Datasets.}
We conduct experiments on seven open-source tabular datasets: \textbf{Microsoft}~\cite{microsoft}, \textbf{YearPrediction}~\cite{yp}, and \textbf{Yahoo}~\cite{yahoo} for regression; \textbf{Forest Cover Type}\footnote{\url{https://www.kaggle.com/c/forest-cover-type-prediction/}}, \textbf{Click}\footnote{\url{https://www.kaggle.com/c/kddcup2012-track2/}}, \textbf{Epsilon}\footnote{\url{https://www.csie.ntu.edu.tw/~cjlin/libsvmtools/datasets/binary.html\#epsilon}}, and \textbf{Cardiovascular Disease}\footnote{\url{https://www.kaggle.com/sulianova/cardiovascular-disease-dataset}} for classification. The details of the datasets are listed as in Table~\ref{tab:data}. Most of the datasets provide train-test splits. For \textbf{Click}, we follow the train-test split provided by the open-source\footnote{Different to the descriptions in the original paper.} of NODE~\cite{NODE}. In all the experiments, we fix the train-test split for fair comparison. For the tasks on learning to rank, we use regression similar to the previous work. For \textbf{Click}, the categorical features were pre-processed with the Leave-One-Out encoder of the scikit-learn library. We used the official validation set of every dataset if it is given. On the datasets that do not provide official validation sets, we stratified to sample $20\%$ of instances from the full training datasets for validation.
\paragraph{Implementation details.}
We implement our various {\ourapproach} architectures with PyTorch on Python 3.7. All the experiments are run on NVIDIA Tesla V100. 
In training, the batch size is 8,192 with the ghost batch size 256 in the ghost batch normalization layers, and the learning rate is initially set to $0.008$ and is decayed by $5\%$ in every 20 epochs.
The optimizer is the QHAdam optimizer~\cite{qhadam} with default configurations except for the weight decay rate $10^{-5}$ and discount factors $(0.8, 1.0)$. For the other methods, the performances are obtained with their specific settings. 
Unlike previous methods requiring carefully setting their hyper-parameters (e.g., NODE~\cite{NODE}), we fix the primary setting of {\ourapproach}s: We set $k_0=5$, $d_0=32$, and $d_1=64$ as default (see Fig.~\ref{fig:framework}(b)). For the datasets with large amounts of raw features (e.g., \textbf{Yahoo} with 699 features and \textbf{Epsilon} with 2K features), we set $k_0=8$, $d_0=48$, and $d_1=96$. We use the dropout rate $0.1$ for all the datasets except for \textbf{Forest Cover Type} without using dropout.
The performances of the other methods are \textbf{hyperparameter-tuned} for best possible results using the Hyperopt library\footnote{\url{https://github.com/hyperopt/hyperopt}} and performed 50 steps of the Tree-structured Parzen Estimator (TPE) optimization algorithm, similar to the settings in~\cite{NODE}. We set the hyper-parameter search spaces and search algorithms of XGBoost~\cite{XGBoost}, CatBoost~\cite{CatBoost}, NODE~\cite{NODE}, and FCNN~\cite{FCNN} as in~\cite{NODE}, while the hyperparameter search settings of Net-DNF~\cite{dnfnet} and TabNet~\cite{tabnet} followed their original papers. The hyperparameters of gcForest~\cite{deepforest} followed its default values. The architectures of FCNN with or without $l^1$-norm regularization were constructed following the FCNN in~\cite{NODE}. The hyper-parameters of these compared methods are selected according to the validation performances, and the performances are obtained on the corresponding test sets.  
\begin{figure*}
    \centering
    \includegraphics[width=0.9\textwidth]{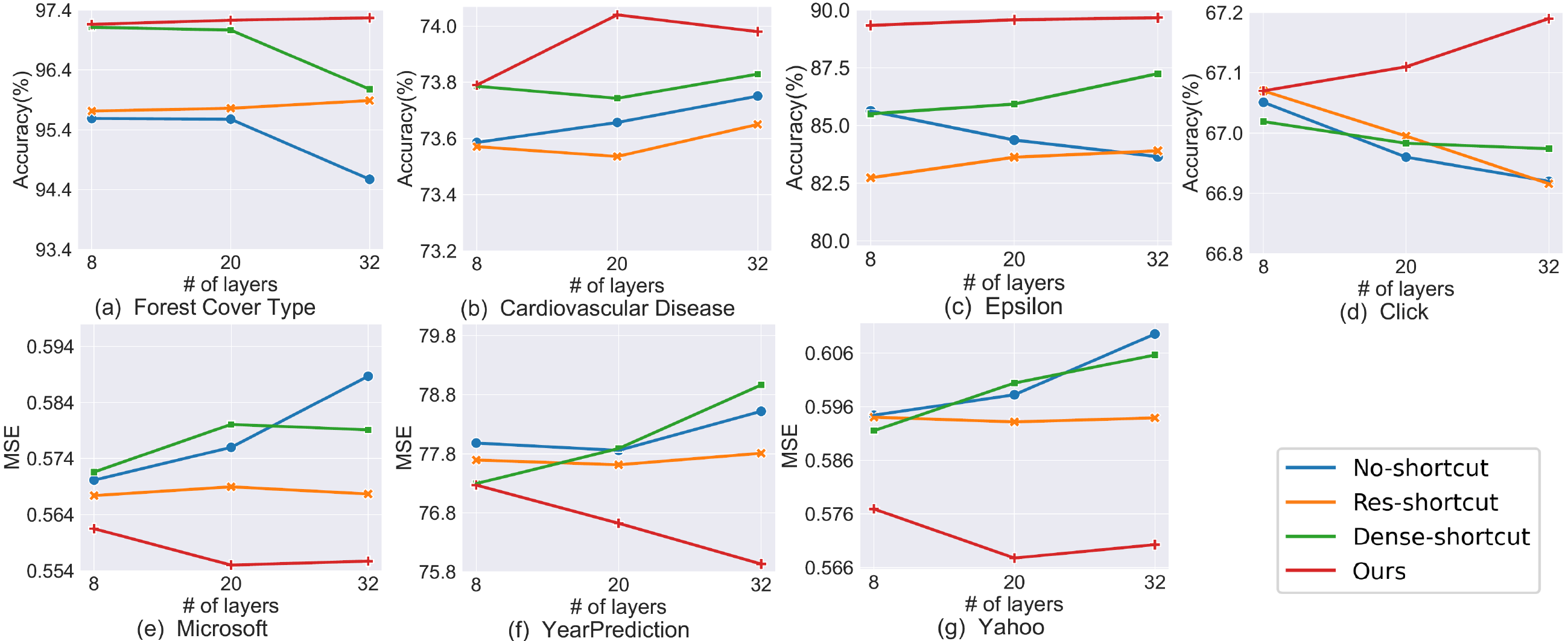}
    % \vskip -0.15 in
    \caption{\textbf{The performances on various datasets with different kinds of shortcuts.} For classification (shown in (a), (b), (c), and (d)), the higher accuracy, the better. For regression (shown in (e), (f), and (g)), the lower MSE, the better. It is obvious that our shortcuts are superior.}
    \label{fig:shortcut}
\end{figure*}
\paragraph{Comparison baselines.}
To evaluate the performances, we compare our {\ourapproach}-20 and {\ourapproach}-32 with several common conventional methods, including XGBoost~\cite{XGBoost}, gcForest~\cite{deepforest}, and CatBoost~\cite{CatBoost}, and the best-known neural networks, including TabNet~\cite{tabnet}, FCNN~\cite{FCNN} with and without the $l^1$-norm regularization, and NODE~\cite{NODE}.
\subsection{Results and Analyses}
\begin{figure}[t]
    \centering
    \includegraphics[width=0.33\textwidth]{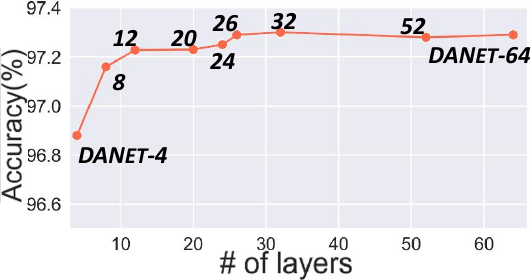}
    \caption{{\ourapproach} performances in different model depths on \textbf{Forest Cover Type}.}\label{fig:deep}
    \vskip -0.8 em
\end{figure}

\begin{figure}[t]
    \centering
    % \vskip -0.15 in
    \includegraphics[width=0.4\textwidth]{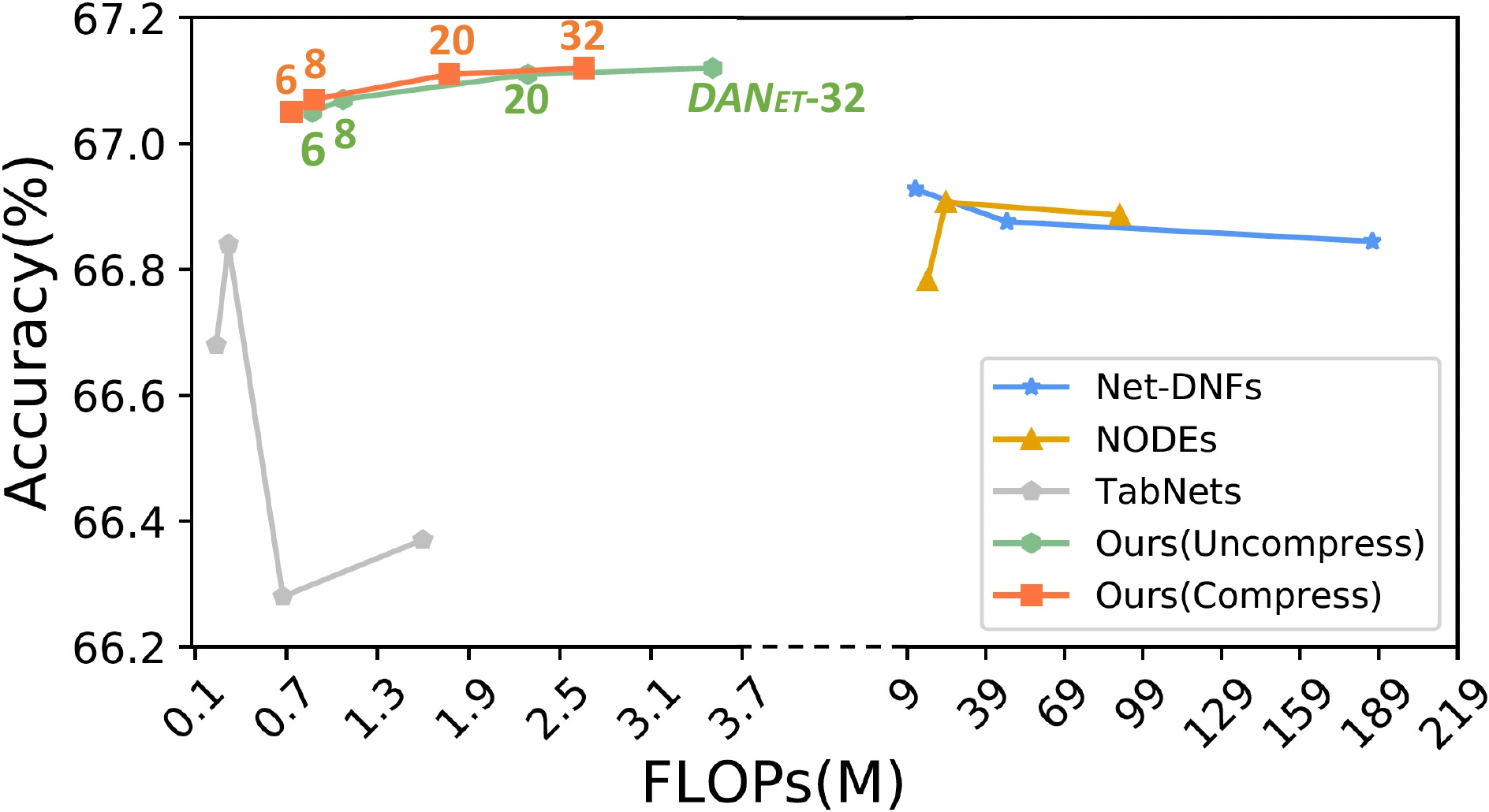}
    % \vskip -0.05 in
    \caption{\small The computational complexity comparison on the \textbf{Click} dataset among {\ourapproach}s and other methods.}\label{fig:flops}
    % \vskip -0.15 in
\end{figure}

\paragraph{Performance comparison.} 
The comparison performances on the seven tabular datasets are reported in Table~\ref{tab:performances}. One can see that our methods (i.e., {\ourapproach}-20 and {\ourapproach}-32) outperform or are comparable with the previous neural networks and GBDTs. Note that the parameters of our {\ourapproach}s are pre-set, while the other methods are specifically \textbf{hyperparameter-tuned} for each dataset. This implies that our {\ourapproach}s are not only better-performing but also easy-to-use. Further, we rank all the methods (except gcForest~\cite{deepforest} and Net-DNF~\cite{dnfnet}, since they can only work on classification) based on the averaged performance ranks on the datasets, and our methods {\ourapproach}-20 and {\ourapproach}-32 attain the best performances among all the methods. Besides, the overall performances of {\ourapproach}-32 are better than {\ourapproach}-20, obtaining performance gain by increasing the model depth.
\paragraph{The effects of shortcuts.}
A key design of our {\ourapproach}s is the special shortcut connections in the basic blocks. To inspect the effects of our proposed shortcuts, we compare {\ourapproach}s with the models with conventional residual shortcuts (Res-shortcut), the models without any shortcuts, and the models with densely connected shortcuts (Dense-shortcut)~\cite{densenet}. For fairness, we only replace our shortcuts with other shortcuts in {\ourapproach}-8, {\ourapproach}-20, and {\ourapproach}-32. The performances are shown in Fig.~\ref{fig:shortcut}. It is evident that {\ourapproach}s with our shortcuts significantly outperform the models with other shortcuts in all the model depth specifications. Besides, one might see that the effects of our proposed shortcuts are more evident in most the cases with deeper {\ourapproach}s. For example, in Fig.~\ref{fig:shortcut}(b), (d), (e), (f), and (g), the performance differences on {\ourapproach}-32 are more noticeable than those on {\ourapproach}-8. This might be because information can be efficiently replenished via our shortcuts, thus helping promote the effectiveness of the deeper models.
\begin{table*}[t]
\caption{\textbf{The mask activation on three synthesized datasets.} Each heatmap has two rows: The top row is for the mask in the main model path, and the bottom row is for the mask in the shortcut path.}
% \vskip -0.15 in
\centering
% \resizebox{0.8\textwidth}{!}{
\begin{tabular}{l|c|c}
\toprule
\multicolumn{1}{c|}{\multirow{2}{*}{Formulas}} & Learn-to-rank & Classification\\ \cmidrule{2-3}
        & \begin{minipage}[b]{0.50\columnwidth}
		\centering
		\raisebox{-.5\height}{\includegraphics[width=\linewidth]{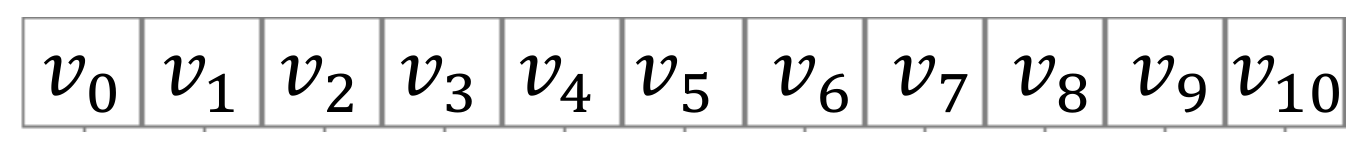}}
	\end{minipage} & \begin{minipage}[b]{0.50\columnwidth}
		\centering
		\raisebox{-.5\height}{\includegraphics[width=\linewidth]{figures/title.pdf}}
	\end{minipage}\\ \midrule
        \circled{1} $y = \sum^5_{i=2} (v^2_i)$ &  \begin{minipage}[b]{0.50\columnwidth}
		\centering
		\raisebox{-.5\height}{\includegraphics[width=\linewidth]{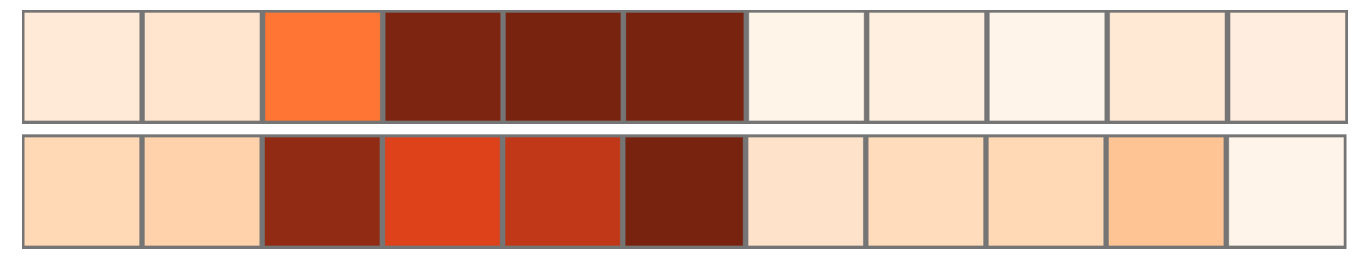}}
	\end{minipage} & \begin{minipage}[b]{0.50\columnwidth}
		\centering
		\raisebox{-.5\height}{\includegraphics[width=\linewidth]{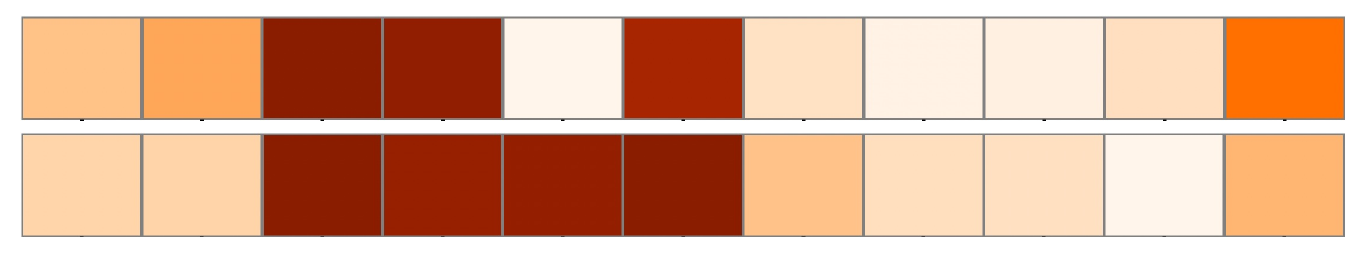}}
	\end{minipage}\\ \midrule
        \circled{2} \small{$y=|\log{|v_0-v_2|} + \cos{(v_5 + \sin{v_6}) - (10^{-8} \times v_{10})}|$} &    \begin{minipage}[b]{0.50\columnwidth}
		\centering
		\raisebox{-.5\height}{\includegraphics[width=\linewidth]{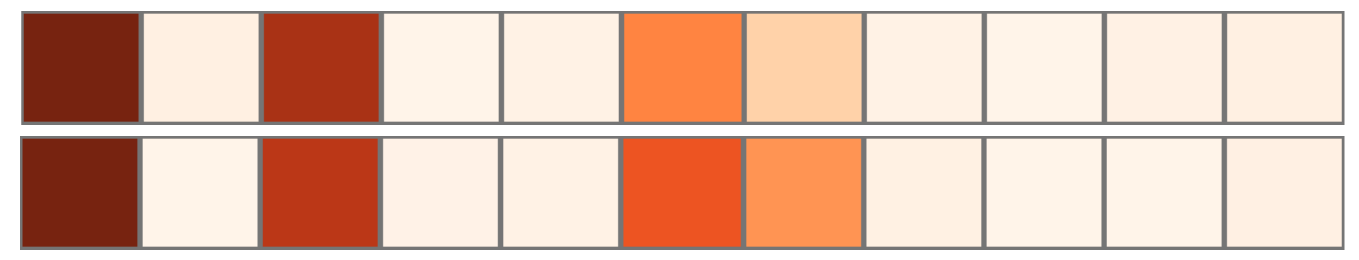}}
	\end{minipage} & \begin{minipage}[b]{0.50\columnwidth}
		\centering
		\raisebox{-.5\height}{\includegraphics[width=\linewidth]{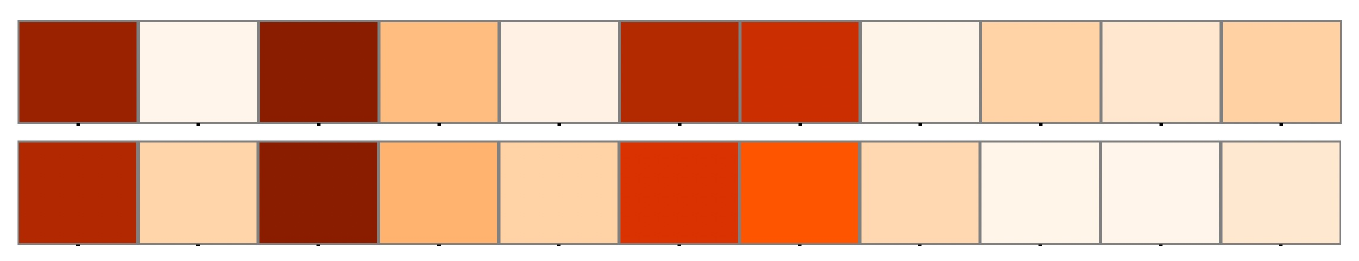}}
	\end{minipage} \\ \midrule
         \circled{3} \small{$y = \sum\limits_{(i,j)\in{\{(6,7), (5, 8)\}}}{-10} \sin{\frac{(v_i + v_j)}{10}} + (v_i + v_j)^2$} &    \begin{minipage}[b]{0.50\columnwidth}
		\centering
		\raisebox{-.5\height}{\includegraphics[width=\linewidth]{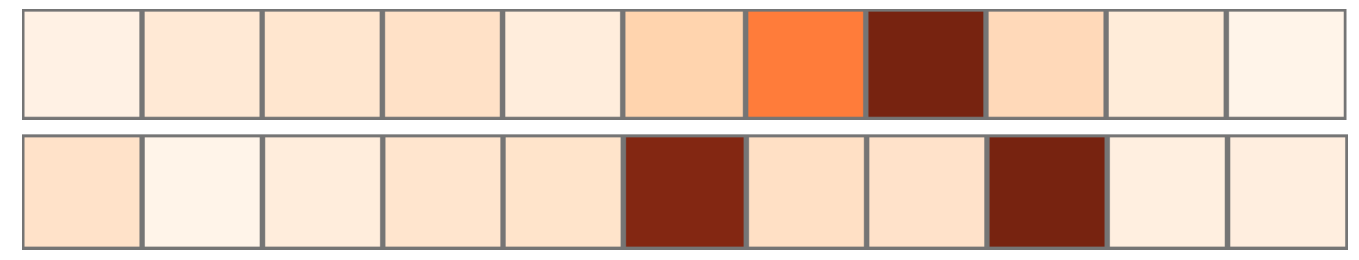}}
	\end{minipage}  & \begin{minipage}[b]{0.50\columnwidth}
		\centering
		\raisebox{-.5\height}{\includegraphics[width=\linewidth]{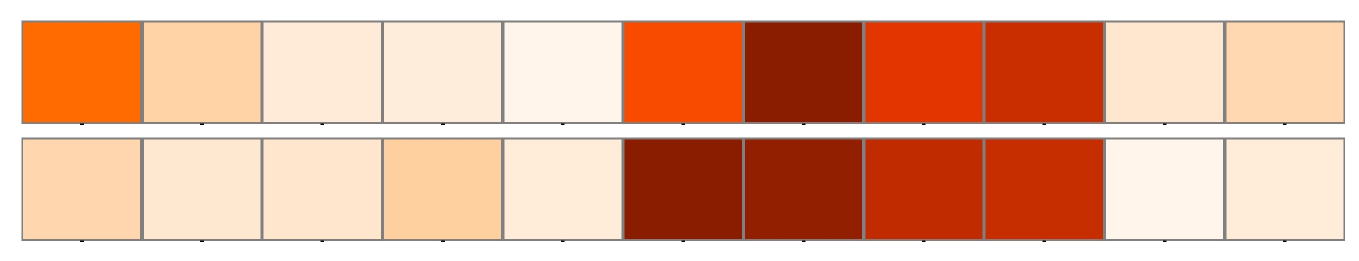}}
	\end{minipage} \\ \midrule
	 \circled{4} \small{$y=$ \circled{1} if $v_1 < 0$; $y=$ \circled{2} if $v_1 > 0$} &    \begin{minipage}[b]{0.50\columnwidth}
		\centering
		\raisebox{-.5\height}{\includegraphics[width=\linewidth]{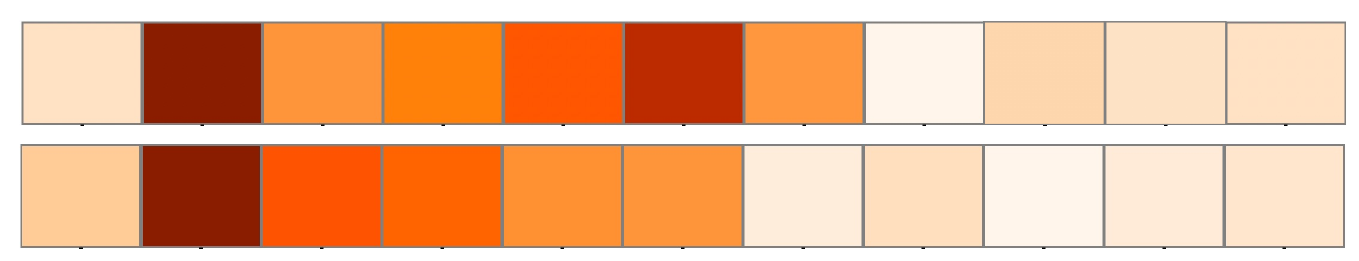}}
	\end{minipage} & \begin{minipage}[b]{0.50\columnwidth}
		\centering
		\raisebox{-.5\height}{\includegraphics[width=\linewidth]{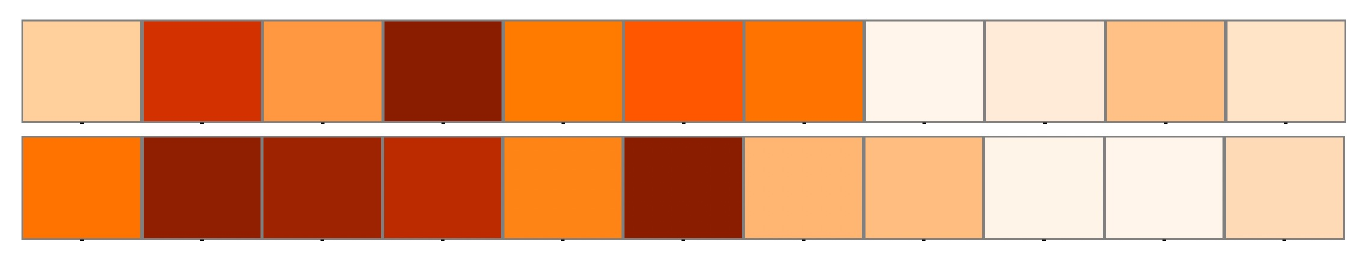}}
	\end{minipage} \\ \bottomrule
\end{tabular}
% }
\label{tab:mask}
\end{table*}
\paragraph{The effects of model depth.}
We show the effects of the {\ourapproach} model depths on the \textbf{Forest Cover Type} dataset in Fig.~\ref{fig:deep}, and we also examined similar phenomena on the other datasets. From Fig.~\ref{fig:deep}, one can see that {\ourapproach}s yield better performances with increasing model depths. However, when {\ourapproach}s get very deep (e.g., deeper than {\ourapproach}-32), the performance gain becomes diminutive. We think this is because tabular data usually have much fewer features than image/text data for very deep networks to exploit. We observe that for {\ourapproach}s, the depths of 20--32 are promising choices.
\paragraph{The effects of model width.}
The number of feature groups, $K$, in an \ourcomponent acts as the model width for \textit{\ourapproach}s. To evaluate the effect of the width $K$, we show the performances of {\ourapproach}-20 with different widths on \textbf{Click} (11 features), \textbf{Forest Cover Type} (54 features), and \textbf{Epsilon} (2K features) in Table~\ref{tab:width}. {\ourapproach}-20 yields considerable performances with width $K=5$. For the datasets with less features (e.g., \textbf{Click} and \textbf{Forest Cover Type}), we only see slight gains with width $K > 5$. For the dataset with more features (\textbf{Epsilon}), $K=8$ seems to be a reasonable choice, which outperforms $K=5$ by $0.13\%$. This may be because a dataset with more features tends to have more feature groups, and thus a larger model width may help in such scenarios.
\paragraph{The effects of sparse masks.}
We inspect the effects of the masks on three synthesized datasets with three different dataset settings. Each dataset contains $7k$ input items with 11 scalar features ($x=\{v_i|i=0, \ldots, 10\}$) generated from an $11$-dimensional Gaussian distribution without feature correlation. Four formulas are used to compute the target $y$ in the first column of Table~\ref{tab:mask}. As for the learn-to-rank tasks, $y$ is used as the prediction targets; as for the classification tasks, $y$ is further transformed into ``0'' or ``1'' using the median of $y$ as the threshold value.
We build an {\ourapproach}-2 with $K=1$, and train it with the synthesized datasets. This model has only one basic block, and there are two masks whose input is the raw features (i.e., the mask of the first \ourcomponent in the main model path and the mask of the \ourcomponent in the shortcut). In this study, we only inspect these two masks after training to convergence, and check whether the mask activation matches the formulas. The mask activation is visualized in Table~\ref{tab:mask}.
\begin{table}[t]
\centering
\caption{{\ourapproach}-20 performances with different widths.}\label{tab:width}
\begin{tabular}{c|ccc}
\toprule
$K$ & Click   & Forest  & Epsilon \\ \midrule
1     & 67.03\% & 96.18\% & 89.13\% \\
5     & 67.11\% & 97.23\% & 89.45\% \\
8     & 67.12\% & 97.21\% & 89.58\% \\
14    & 67.15\% & 97.22\% & 89.61\% \\
20    & 67.15\% & 97.23\% & 89.63\% \\ \bottomrule
\end{tabular}
\end{table}
Taking the learn-to-rank tasks as example, our first question is: \textit{Can our masks distinguish the target-relevant and target-irrelevant features?} For Formula \circled{1}, one can see that only the features $v_2, v_3, v_4$, and $v_5$ are target-relevant, and the corresponding values in the masks are highly responding to them. Similar results can be seen in the other cases. Especially, we introduce a term with regard to $v_{10}$ tending to zero in Formula \circled{2}, and one can see that the masks do not respond to it, which shows that our proposed mask is data-driven and robust. Our second question is: \textit{Can our masks group correlative features?} In Formula \circled{2}, we can regard $v_0$ and $v_2$ as in one group, and $v_5$ and $v_6$ as in another group. We can see that in the masks, the values representing $v_0$ and $v_2$ have close values, and so do $v_5$ and $v_6$. In Formula \circled{3}, there are two feature groups: $(v_6,v_7)$ and $(v_5, v_8)$. Correspondingly, a mask ``selects'' $v_6$ and $v_7$, and the other one ``selects'' $v_5$ and $v_8$. As for the piecewise function in Formula \circled{4}, $v_1$ (as a condition) and all the features used in Formulas \circled{1} and \circled{2} are considered by the masks. In summary, one can see that our proposed masks not only can find target-relevant features, but also have the ability to dig out feature relations. Similar conclusions can be drawn for the classification tasks.
\paragraph{Computational complexity comparison.} We compare the computational complexities in the inference phases of {\ourapproach}s with the performance-competitive neural networks, TabNet, NODE~\cite{NODE}, and Net-DNF~\cite{dnfnet} (see Fig.~\ref{fig:flops})\footnote{The hyperparameters of the four compared TabNets are: [$\lambda_{sparse}$, $N_d$, $N_a$, $N_{steps}$, $B_V$, $m_B$] = [1e-4, 32, 32, 3, 256, 0.9], [1e-4, 32, 64, 5, 256, 0.9], [1e-4, 32, 64, 7, 256, 0.9], [1e-4, 64, 64, 10, 256, 0.9]; NODEs: [number of layers, total number of trees, tree depth, output dimension of trees] =  [2, 1024, 6, 3], [4, 1024, 6, 3], [4, 2048, 6, 3]; Net-DNFs: [number of formulas, feature selection beta] = [512, 1.0], [1024, 1.3], [2048, 1.6].}. The FLOPS of ensemble learning based methods (i.e., NODE and Net-DNF) are generally several times those of {\ourapproach}s and TabNet. Besides, it is obvious that, under some identical complexities, our models are often the best-performed ones. Seeing the grey curve in Fig.~\ref{fig:flops}, TabNet cannot obtain the performance gains when keeping enlarging the model size, which are not very extensible compared to ours. After model compression by the structure re-parameterization performed on {\ourcomponent}s, the FLOPS of our {\ourapproach}s are reduced by $14.8\%$--$23.0\%$ (compared the red and green curves). As for one single \ourcomponent, the FLOPS are reduced by $49.02\%$ with the input and output feature sizes of 32.
\section{Conclusions}
In this paper, we proposed a family of parse-tree-like deep neural networks, {\ourapproach}s, for tabular learning. We designed a novel neural component, \ourcomponent, for tabular data, which automatically selects correlative features and abstracts higher-level features from the grouped features. We also provided a structure re-parameterization method which can largely reduce the computational complexity of \ourcomponent. We developed a basic block based on {\ourcomponent}s, and {\ourapproach}s in various depths were built by stacking such blocks. A special shortcut in the basic block was introduced, increasing the diversity of feature groups. Experiments on several public datasets verified that our {\ourapproach}s are effective and efficient in processing tabular data, for both classification and learn-to-rank tasks. Besides, using synthesized datasets, we show that the proposed masks can find feature correlations. Besides, the ablation studies explored the effectiveness of model depths and widths, which suggested that a wider and deeper \ourapproach is beneficial but the extreme depth and width architectures were not recommended due to the limited spaces of tabular features.
\section*{Acknowledgment}
This research was partially supported by National Key R\&D Program of China under grant No. 2018AAA0102102, National Natural Science Foundation of China under grants No. 62176231 and 62106218, Zhejiang public welfare technology research project under grant No. LGF20F020013, Wenzhou Bureau of Science and Technology of China under grant No. Y2020082. 
Yao Wan was supported in part by National Natural Science Foundation of China under grand No. 62102157.
D. Z. Chen was supported in part by NSF Grant CCF-1617735. 
% We would like to thank all the anonymous reviewers, SPCs and ACs for their suggestive comments and feedback.
\bibliography{ref}
\end{document}